\pdfoutput=1

\documentclass[10pt,twocolumn,letterpaper]{article}

\usepackage{cvpr}              

\usepackage{graphicx}
\usepackage{amsmath}
\usepackage{amssymb}
\usepackage{booktabs}

\usepackage[pagebackref,breaklinks,colorlinks]{hyperref}

\usepackage[capitalize]{cleveref}
\crefname{section}{Sec.}{Secs.}
\Crefname{section}{Section}{Sections}
\Crefname{table}{Table}{Tables}
\crefname{table}{Tab.}{Tabs.}

\def\cvprPaperID{*****} 
\def\confName{CVPR}
\def\confYear{2023}

\begin{document}

\title{\ Mutilmodal Feature Extraction and Attention-based Fusion for Emotion Estimation in Videos}

\author{Tao Shu,Xinke Wang,Ruotong Wang,Chuang Chen\\
AHU-IAI AI Joint Laboratory, Anhui University\\
Institute of Artificial Intelligence, Hefei Comprehensive National Science Center\\
{\tt\small {wa21201011, wa22101009,wa21301031,wa21301027}@stu.ahu.edu.cn}
\and
Yixin Zhang, Xiao Sun\thanks{Corresponding author.}\\
Institute of Artificial Intelligence, Hefei Comprehensive National Science Center\\
{\tt\small zhyx12@ustc.edu.cn, sunx@iai.ustc.edu.cn}
}

\maketitle

\begin{abstract}
The continuous improvement of human-computer interaction technology makes it possible to compute emotions. In this paper, we introduce our submission to the CVPR 2023 Competition on Affective Behavior Analysis in-the-wild (ABAW).
Sentiment analysis in human-computer interaction should, as far as possible Start with multiple dimensions, fill in the single imperfect emotion channel, and finally determine the emotion tendency by fitting multiple results. Therefore, We exploited multimodal features extracted from video of different lengths from the competition dataset, including audio, pose and images. Well-informed emotion representations drive us to propose a Attention-based multimodal framework for emotion estimation. Our system achieves the performance of 0.361 on the validation dataset. The code is available at [https://github.com/xkwangcn/ABAW-5th-RT-IAI].
\end{abstract}

\section{introduction}
\label{sec:intro}
In human-computer interaction, computers are required to capture crucial information, perceive changes in human emotions, form expectations, make adjustments, and respond to what we call affective computing. With the demand of consumer scenarios and industry applications pulling, affective computing is also becoming increasingly widely used in real life, and has been integrated into various aspects of production life such as online education, critical care, fatigue driving detection, personalized recommendation, and has application prospects in future fields such as virtual reality. Numerous relevant datasets exist to support theoretical studies, such as AffectNet, Aff-wild2, RAFDB.

\begin{figure*}
    \centering
    \includegraphics[width=0.9\linewidth]{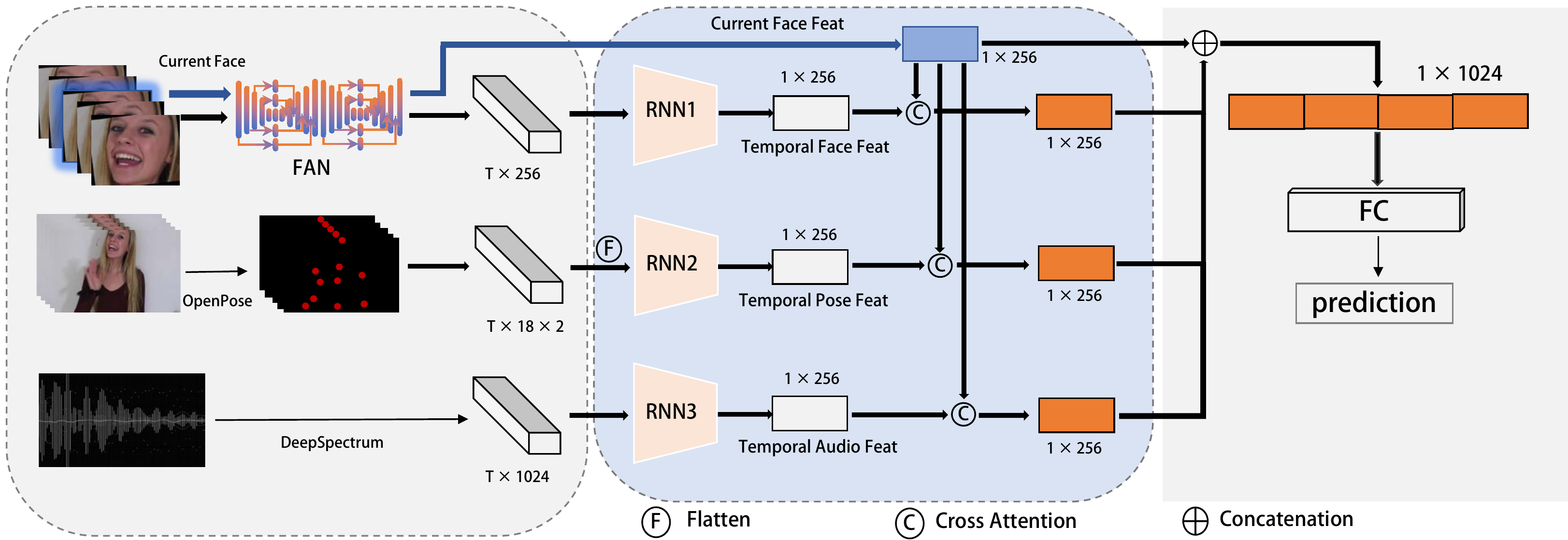}
    \caption{Framework of our proposed Mutil Modal Attention. 
    The picture sequence of the face of the face of the currently predicted frame will be extracted by FAN and composed of feature sequences, the feature sequence and the face sequence are of equal length, and the visual face features of the current frame will be taken out from the corresponding position in the feature sequence for later fusion. The face feature sequence, pose information sequence and audio feature sequence will go through three two-layer RNNs respectively to obtain features in time dimensions, which will be Attention calculated with the Current Face Feature respectively, so as to obtain three different face feature attention, and finally we directly concatent these attention features and connect the FC layer to predict the expression.
    }
    \label{fig:overrall-model}
    \vspace{-0.4cm}
\end{figure*}

\section{Related Work}
\label{sec:related work}

\subsection{Image-based facial expression recognition}
Image-based facial expression recognition has been extensively studied in recent years. Generally, a FER system mainly consists of three stages, namely face detection, feature extraction, and expression recognition. Traditionally, the hand-crafted features were developed to describe different facial expressions, such as LBP \cite{01-shan2009facial}, HOG \cite{02-dalal2005histograms} , and SIFT \cite{03-ng2003sift}. However, these features lack of generalization ability under some challenging scenarios, such as poor illumination conditions.

Later, many in-the-wild facial expression databases were developed that enabled the research of FER in more challenging environments. Recently, deep learning has greatly improved the FER research \cite{04-ding2020occlusion,05-ding2017facenet2expnet,06-hasani2020breg,07-kim2015hierarchical}. A region attention network is proposed in \cite{08-wang2020region} to adaptively capture the importance of facial regions for occlusion and pose variant FER. Self-Cure network \cite{09-wang2020suppressing} was proposed that suppresses the uncertainties caused by ambiguous expressions and the suggestiveness of the annotators.

\subsection{Audio-based emotion estimation}
For audio modality, several features are widely used in affective computing. It usually can get a good performance when combining with other modality features. There are many kinds of features can be extracted from audio. Such as eGeMAPS \cite{10-eyben2015geneva} and DeepSpectrum \cite{11-amiriparian2017snore}.  eGeMAPS is an extension of GeMAPS, it added some extended features based on GeMAPS. DeepSpectrum is the method you can get deep features of audio signals, which is based on a pre-trained image recognition Convolutional Neural Networks (CNNs).

\subsection{Pose-based emotion estimation}
Gait emotion recognition is to mine deep information from the human gait and model human emotion into discrete space: one-dimensional space consisting of happy, sad, neutral, angry, etc. or continuous space: three-dimensional space formed by pleasure, arousal, and dominance \cite{12-mehrabian1996pleasure}.
Previous work \cite{13-venture2014recognizing,14-crenn2016body} has used psychologist-validated, hand-crafted feature to classify emotions, which has the disadvantage of being cumbersome and not allowing for the understanding of higher-level emotion-related feature. The current work is mainly based on deep learning methods, which can be divided into three categories. The first class, represented by \cite{15-randhavane2019identifying}, uses the LSTM's expressive power over sequences for depth feature extraction, while fusing the hand-crafted feature for emotional classification using a random forest classifier. The second category is represented by \cite{16-narayanan2020proxemo}, which exploits the ability of CNN to express image feature for feature extraction of images composed of coordinates of human motion joints. It is worth noting that this paper focuses on multi-view gait images to increase the robustness of the model. The third category considers that joints linked according to the human biological chain can be considered as a graph structure in a non-Eulerian space, which can be processed using the GCN architecture. \cite{17-bhattacharya2020step} achieves optimal performance by using ST-GCN \cite{18-yan2018spatial} for feature extraction of gait skeleton data, and also the Encoder-Decoder architecture is used in this paper for the synthesis of gait samples.

\setlength{\tabcolsep}{6pt}
\begin{table*}[t]
\centering
\begin{tabular}{l|cccccccc|ll}
    \toprule[1pt]
    Val Set & \ Neutral & \ Anger & \ Disgust & \ Fear & \ Happiness & \ Sadness & \ Surprise & \ Other & \ Acc & \ Avg(F1). \\ 
    \midrule
    Official  & 52 & 6 & 21 & 28 & 49 & 54 & 17 & 43 & 45 & 33.7\\
    Split1  & 50 & 17 & 9 & 24 & 49 & 47 & 23 & 39 & 43 & 32.2 \\
    Split2  & 59 & 12 & 1 & 1 & 49 & 25 & 33 & 64 & 53 & 30.0 \\
    Split3  & 64 & 9 & 8 & 6 & 58 & 23 & 18 & 63 & 55 & 31.0 \\
    Split4  & 62 & 28 & 19 & 9 & 44 & 52 & 10 & 44  & 48 & 33.6 \\
    Split5  & 53 & 27 & 16 & 18 & 40 & 51 & 30 & 49 & 45 & \textbf{35.5} \\
    \bottomrule[1pt]
\end{tabular}
\caption{Comparison of the expression F1 scores (in \%) between different Split}
\label{tab:fan_on_val}
\end{table*}

\setlength{\tabcolsep}{4pt}
\begin{table*}[t]
\centering
\begin{tabular}{l|cccccccc|ll}
    \toprule[1pt]
    Modal & \ Neutral & \ Anger & \ Disgust & \ Fear & \ Happiness & \ Sadness & \ Surprise & \ Other & \ Acc & \ Avg(F1). \\ 
    \midrule
    current face  & 52 & 6 & 21 & 28 & 49 & 54 & 17 & 43 & 45 & 33.7\\
    only video  & 1 & 23 & 5 & 18 & 1 & 75 & 54 & 61 & 53 & 29.5 \\
    concat fusion    & 58 & 23 & 9 & 21 & 42 & 53 & 33 & 52 & 48 & 31.8 \\
    \textbf{attention fusion}  & 58 & 32 & 11 & 16 & 34 & 51 & 28 & 59 & \textbf{49.2} & \textbf{36.1} \\
    \bottomrule[1pt]
\end{tabular}
\caption{Comparison of the expression F1 scores (in \%) between different modality and fusion on validation}
\label{tab:different_modal}
\end{table*}

\section{Approach}
\label{sec:Approach}
Noting that the pose information of the characters in the dataset is rich, we add the pose change feature to this task. In this work, We propose a method to enhance the attention to the features of the current frame based on the temporal features of the face picture sequence, pose and audio in the current frame context for expression recognition.
The overall framework of our approach is shown in the Figure \ref{fig:overrall-model}, The picture sequence of the face of the face of the currently predicted frame will be extracted by FAN and composed of feature sequences, the feature sequence and the face sequence are of equal length, and the visual face features of the current frame will be taken out from the corresponding position in the feature sequence for later fusion. The face feature sequence, pose information sequence and audio feature sequence will go through three two-layer RNNs respectively to obtain features in time dimensions, which will be Attention calculated with the Current Face Feature respectively, so as to obtain three different face feature attention, and finally we directly concat these attention features and connect the FC layer to predict the expression.

\subsection{Multimodal feature extraction}
{\bf Facial features}
In this paper, the facial feature is extracted by FAN, this is a single, end-to-end model that utilizes deep learning methods for facial sentiment analysis under natural conditions. We have pretrained FAN on Affect-Net Dataset and finetuned on 
Aff-Wild2 Datasets, in the finetune stage, considering the efficiency of computing power, we selected a face picture every 5 frames in 30 frames to form a data set, which made the calculation amount one-fifth of the original, than we use the FAN to extract facial feature from single face image. 

In the later fusion stage, we combine the facial features extracted by FAN according to the order of the images, then input them to a two-layer RNN to extract features in the time dimension.

{\bf Audio features}
We extracting three audio features  ComParE2016, eGeMAPS, DeepSpectrum. But due to the time is limited, we only choose the DeepSpectrum to finish this challenge. We using DenseNet121 network and setting windows size to 1 second and hop-size to 500 ms, then get a 1024-dimensional feature vector. For the EXPR task, the way is directly input the extracted features with other modalities to our overall model.

{\bf Dynamic pose features}
For the Dynamic pose features, we first used Openpose to extract the 2D coordinates of the skeletal points of the characters in the original video screen, and the frequency of extraction was detect one frame every 0.5 second. So that it has same frequency as face and audio

\subsection{Attention-based fusion module}
We have made Cross Attention to the temporal feature information for each modality with the current frame. These attention features can misread the targeted extraction of the context information to the face features of the current face. Then the current face and three attention feature are concatenated together and fed into a FC layer to predict expression.

\section{Experiments}

\subsection{DataSets}
The datasets we used is Aff-Wild2 \cite{19-kollias2022abaw,20-kollias2022abaw,21-kollias2021distribution,22-kollias2021analysing,23-kollias2021affect,24-kollias2020analysing,25-kollias2019expression,26-kollias2019face,27-kollias2019deep,28-zafeiriou2017aff}, which provided by the fifth Affective Behavior Analysis in-the-wild (ABAW) Competition\cite{29-kollias2023abaw}. It contains 598 videos, and including three tasks. For the Expression Classification Challenge, the ABAW competition provide 247 of them as train datasets, 70 of them as validation datasets, and 228 of them as the test datasets. We found one file named 122-60-1920x1080-2.txt is appeared in both train datasets and validation datasets, so we delete it in the validation datasets. For improve our model's generalization and robustness performance, we also re-split the train datasets and validation randomly to 5 folds. We combine 4 folds of these for train and last one folds for validation so we have create 5 kinds of splits. We do the experiments both on official split and our own split.

For different modalities, we did some down-sampled work to reduce train time. For image modality, we down-sampled these video frames with 5 times to reduce the train images, used one image each 5 frames. And we delete the frame with invalid annotation. It's about 180000 images are used. For audio and pose features, we extracted two pieces of data each 1 second. 

\subsection{Experimental Settings}
This part we will introduce some details in our experiments. We trained the models on one RTX3090.

{\bf Pretrain FAN}
We pretrained the FAN on Affect-Net. The implementation was done using open-source software, specifically PyTorch for the deep learning part. We trained the networks using Adam optimizer with a decrease of the learning rate by 10 every 15 epochs. All the hyper-parameters were validated using a randomized grid search. In particular, we validated the weight decay in the range [0.0, 0.01], the learning rate in the range [0.0001; 0.01] and the optimizer’s parameters beta1 and beta2 in the range [0.0; 0.999]. Additional details and specifications are provided in the Supplementary Information.

{\bf Finetune FAN}
When we finetune the FAN on Aff-Wild2 at most 30 epochs. We trained the networks using AdamW optimizer and we set the weight decay to 0.05. We set the backbone learning rate to 4e-5 and downstream predictor learning rate to 4e-3. These learning rate will multiply 0.5 when validation F1 score can not decline for 2 epochs. As shown in \cref{tab:fan_on_val}, we compared the expression F1 scores (in \%) between different Split. We find the FAN achieves the best score on average in the 15th epoch.

{\bf Fusion Training}
In order to get the face,audio and pose sequence, we use the window size of 6 seconds and select 2 frames per second, that means T of the input of network is 12.

To train the multi modal fusion network, we have froze parameters of FAN and take it as an extractor of facial feature. For the other parameters except FAN, we set the learning rate to 0.02, and will multiply 0.5 when validation F1 score can not decline for 2 epochs. Because of limitation of memory for the RTX3090, we set the batchsize to 4.

\subsection{Results}
As shown in \cref{tab:different_modal}, we did five kinds of experiments for single modality and multi modalities. In details, including current face, only video, cancat fusion and attention fusion. For cancat fusion, it means we only cancat the three temporal modality features and then do the predictions directly. For attention fusion, as shown in Figure1, we do the attention action for the above three modalities with face modality, and then concat all of them. The results on the validation set is shown on Table 2. Based the results we can know that the attention fusion performance better than other four modalities.

\section{Conclusion}
In this paper, We propose a Attention-based multimodal framework for feature fusion to emotion estimation, including audio, pose and images. Our system achieves the performance of 0.361 on the validation dataset

{\small
\bibliographystyle{ieee_fullname}
\bibliography{egbib}
}

\end{document}